\documentclass[letterpaper, paper,11pt]{AAS}		

\usepackage{bm}
\usepackage{amsmath}
\usepackage{amssymb}
\usepackage{subfigure}
\usepackage{mathtools}
\usepackage[colorlinks=true, pdfstartview=FitV, linkcolor=black, citecolor= black, urlcolor= black]{hyperref}
\usepackage{overcite}
\usepackage{footnpag}			      	
\usepackage[many]{tcolorbox} 
\usepackage[subtle]{savetrees}

\PaperNumber{25-295}

\newtcolorbox{boxA}{
    boxrule = 1.5pt,
    colframe = black 
}

\begin{document}

\title{Adaptive Navigation Strategy for Low-Thrust Proximity Operations in Circular Relative Orbit}

\author{D. Ruggiero\thanks{Research Fellow, Department of Mechanical and Aerospace Engineering, Politecnico di Torino, C.so Duca degli Abruzzi 29, Torino, Torino, 10129, Italy. (dario.ruggiero@polito.it)},  
M. Mancini \thanks{Assistant Professor, Department of Mechanical and Aerospace Engineering, Politecnico di Torino, C.so Duca degli Abruzzi 29, Torino, Torino, 10129, Italy. (mauro.mancini@polito.it)},
\ and E. Capello \thanks{Full Professor, Department of Mechanical and Aerospace Engineering, Politecnico di Torino, C.so Duca degli Abruzzi 29, Torino, Torino, 10129, Italy, and with CNR-IEIIT, Politecnico di Torino, C.so Duca degli Abruzzi 29, Torino, Torino, 10129, Italy. (elisa.capello@polito.it)}
}


\maketitle{}

\begin{abstract}
This paper presents an adaptive observer-based navigation strategy for spacecraft in Circular Relative Orbit (CRO) scenarios, addressing challenges in proximity operations like formation flight and uncooperative target inspection. The proposed method adjusts observer gains based on the estimated state to achieve fast convergence and low noise sensitivity in state estimation. A Lyapunov-based analysis ensures stability and accuracy, while simulations using vision-based sensor data validate the approach under realistic conditions. Compared to classical observers with time-invariant gains, the proposed method enhances trajectory tracking precision and reduces control input switching, making it a promising solution for autonomous spacecraft localization and control.
\end{abstract}

\begin{boxA}
    This document is an author’s original manuscript (pre-print) of the paper AAS 25-295 presented at the 35th AAS/AIAA Space Flight Mechanics Meeting, Kaua'i, Hawaii, January 19-23, 2025.
\end{boxA}
\section{Introduction}

Autonomous space operations are advancing rapidly, enabling new capabilities in deep space exploration, satellite maintenance, and planetary missions, but also introducing challenges. 
In particular, a crucial issue is the realisation of autonomous Guidance Navigation and Control (GNC) systems for managing the motion of the spacecraft in real-time without human intervention, mainly due to communication delays with ground stations. 
Specifically, the guidance function consists of determining the desired motion of the spacecraft so as to provide the controller with the reference to follow.  
Indeed, the control function refers to the algorithms that, based on both the actual and desired motion of the spacecraft, calculate the control forces (physically applied by the actuators) required to follow the guidance output. 
Finally, autonomous navigation function must combine the outputs of onboard sensors (such as star trackers, optical cameras, and IMUs) within suitable navigation algorithms to determine position and velocity of the spacecraft without relying on Earth-based signals like GPS. 
For the success of future deep space missions, it is necessary to ensure the accuracy, robustness and reliability of these systems, which are required to operate autonomously in harsh conditions with environmental disturbances, space debris and power constraints. \cite{di2018survey,quanz2022large} 
In the context of autonomous space missions, proximity formation flight and cooperative (or non-cooperative) inspection of targets pose particular challenges, especially with regard to real-time localization of spacecraft. 
Indeed, spacecraft flying in proximity formation must accurately coordinate their movements in real time to avoid collisions whilst meeting mission requirements. This imposes stringent constraints on autonomous navigation systems, which are required to manage a multitude of real-time information. \cite{ito2024formation,hansen2020linear,matsuo2022high} 
Also, uncooperative target inspection involves observing or approaching objects that are not transmitting their position or cooperating with the inspection process, such as defunct satellites or space debris. This task requires highly sophisticated onboard algorithms to autonomously interpret sensor data, model the target's movement, and maintain a safe distance while collecting detailed information.\cite{FLORESABAD20141} Both proximity formation flight and uncooperative target inspection require overcoming challenges like limited computational resources, communication delays, and the unpredictability of space environments, making them fundamental research topics for the advancement of autonomous space operations.

In order to study a scenario that is representative of both space missions described above, this paper proposes an autonomous GNC strategy to force a satellite (chaser) to move around a target at a fixed distance, i.e. the chaser has to move in a Circular Relative Orbit (CRO) around the target.  CRO trajectories are derived from the Restricted Two-body Problem, as solution of the Clohessy-Wiltshire Equations for unperturbed orbital relative motion. \cite{alfriend2009spacecraft}
In this work, the desired CRO trajectory is generated through a guidance algorithm based on the Lyapunov Guidance Vector Field (LGVF) method. Previous works has already demonstrated the validity of this guidance strategy for tracking CRO trajectory in low-thrust formation flight missions.\cite{ruggiero2024,sarvadon2024high}
The LGVF approach is used to generate a velocity field based on spacecraft relative position, that allows to track the desired CRO trajectory. 
Consequently, the control problem is formulated as a velocity tracking problem, and it is addressed by a velocity feedback controller. 
However, the performance of both guidance and control strategies are heavily affected by the estimation accuracy of the spacecraft position and velocity.   
In fact, noisy signals lead to low CRO tracking accuracy and high control input switching frequency, increasing fuel consumption and the risk of not meeting mission requirements. For this reason, the effect of sensors noise and estimation accuracy are investigated in this paper.
Wide range of different sensors can be used for spacecraft relative position measurements, such as vision-based cameras, lidar, radio frequency and laser.\cite{zhang2022survey} 
In this study, we consider on-board vision-based cameras for measuring the relative position of the satellite, thus addressing the noise of the signals according to the performance typically offered by these devices. Vision-based cameras often rely on computer vision and image processing to determine the relative position of another spacecraft, and recent studies show high capabilities of visual navigation algorithms, allowing to reach cm-pose estimation accuracy with reasonable computational effort. \cite{cassinis2019review,kaki2023real,pesce2019autonomous,tweddle2015relative}


Then, the main scope of this study is to implement and to compare different observer-based navigation strategies for a spacecraft in CRO around a target. In the proposed strategies, the relative position is measured through the on-board sensors and is taken in input by the state observer which reconstructs the full state of the spacecraft (relative position and velocity). In this work, different state observers (Luenberger, Kalman filter, Sliding Mode) are designed considering the relative motion  dynamics of the spacecraft and the measurement noise produced by realistic vision-based sensors. 
Luenberger observer (LO) is a well known state estimation technique which  relies on the system's mathematical model and feedback correction to guarantee asymptotic convergence of the estimation errors.  \cite{1099826} However, model uncertainties, parameter variations and signal noises can strongly compromise the convergence properties of the LO. 
Sliding Mode Observers (SMO) are state estimation techniques highly suitable for nonlinear systems, in which the estimation error is forced to "slide" along a surface. These observers exhibit high robustness against model mismatches and external perturbations, while showing attractive noise resilience properties. \cite{doi:10.1080/0020717031000099029,ruggiero2023design,ruggiero2023failure}
However, both LO and SMO struggle to combine low noise sensitivity with fast convergence properties. Indeed, high observer gains ensure quick convergence of the observer values to the system states, but also cause increased sensitivity to noise. On the other side, observers with low gains exhibit reduced sensitivity to noise, but require long time to converge for large estimation error. 

Recent studies have explored time-varying observer gains, ensuring convergence of estimation under specific conditions.\cite{moreno2017levant}
In this work, a suitable adaptive law for observer gains is proposed, in order to achieve fast convergence and low noise sensitivity of the observer values. The adaptive law is based on the estimated position error, with specific consideration of spacecraft motion under the CRO trajectory conditions. The core idea is to implement a reactive observer when the trajectory error is large, while increasing estimation accuracy as the system approaches the desired trajectory.
By reducing the estimation error as the system converges to the desired trajectory, the feedback gain is decreased to reduce the sensitivity to noise. This ensures that the closed-loop system achieves high tracking accuracy and smooth control action. 
The effectiveness of the proposed  strategy is validated through both Lyapunov-based stability analysis and numerical simulations, taking into account realistic sensor noise and actuator characteristics. The results show that, compared to constant-gains observers, the proposed adaptive observer law effectively enhances tracking accuracy, minimizes control input chattering, and ensures robust system performance, making it a promising solution for spacecraft trajectory control. 

The rest of the article is organized as follows. Section "DYNAMICS MODELS" explains the equations ruling the relative motion of the chaser with respect to the target and gives the solution identifying the CRO trajectory. Then, the proposed GNC algorithms, together with the adaptive law realizing time-varying gains, is detailed in the section "GNC STRATEGY". The simulation scenario and results of the numerical simulations are enclosed in the section "SIMULATIONS", while some concluding remarks are in the "CONCLUSIONS".

\section{Dynamics Models}
\label{sc:model}
This section details the mathematical models to describe the motion of the spacecraft relative to the target. Equation of motion are derived from the propagation of the relative dynamics, usually known as Hill’s or Clohessy-Wiltshire (CW) Equation, and it is expressed in the Local-Vertical-Local-Horizontal (LVLH) reference frame.\cite{clohessy1960terminal} The LVLH reference frame is centered in the target spacecraft orbiting the Earth on a circular orbit of radius $r_o$, with the x-axis oriented as the orbital velocity direction, z-axis pointing to the center of the Earth, and y-axis completing the terns. Traslational dynamics is expressed as
\begin{align}
        \ddot{x}&=2\omega\dot{z}+u_x+w_{d,x}\nonumber,\\
        \ddot{y}&=-\omega^2y + u_y+w_{d,y},\label{eq:hill}\\
        \ddot{z}&=3\omega^2z-2\omega\dot{x}+ u_z+w_{d,z},\nonumber
\end{align}
where $\omega=\sqrt{\mu/r_o^3}$ is the orbital angular rate, $p=[x,\ {y},\ {z}]^T$ and $v=[\dot{x},\ \dot{y},\ \dot{z}]^T$ are spacecraft position and velocity, $u=[u_x,\ u_y,\ u_z]^T$ is the control input, while external disturbances and unmodelled dynamics are taken into account in $w=[w_{d,x},\ w_{d,y},\ w_{d,z}]^T$. For LEO analysis, external disturbances are mainly related to atmospheric drag and J2 effect. A simplified disturbance model is included, and given as constant acceleration of the order of magnitude $d_{ex}\sim-10^{-7}\ $m/s$^2$, and acting on the x-axis. Process noise is included in $w$, and it is modelled using normal distribution with variance $\sigma_{pn}=10^{-8}$.

Vision-based relative navigation is considered, providing relative position measurements with $cm$ accuracy (during proximity operations). Sensors measurements are modelled including output noise as normal distribution with variance $\sigma_y=10^{-2}$.  
Low-thrust continuous thrusters are considered. Actuators model is given by a first order filter providing a maximum acceleration of $10^{-5}\ $m/s$^2$.

In this paper, we are mainly focusing on a particular solution of Equation \eqref{eq:hill} for unperturbed motion ($w=0$). The proposed solution is expressed as the CRO trajectory, and it is given as
\begin{align}
        {x}(t)&=-R\cos{(\omega t + \phi)},\nonumber\\
        {y}(t)&=\sqrt{3}R\sin{(\omega t + \phi)}/2\label{eq:CRO},\\
        {z}(t)&=R\sin{(\omega t + \phi)}/2,\nonumber
\end{align}
where $\phi\geq0$ is a phase angle, and $R>0$ is the CRO trajectory radius. The combination of Equations \eqref{eq:hill} and \eqref{eq:CRO}, with $u=w=0$, leads to $\ddot{x}=0$. The CRO trajectory is studied as low-cost trajectory for missions where the spacecraft is required to moves around the target at a fixed distance (i.e., formation flight missions\cite{danzmann2000lisa,ando2010decigo}, non-cooperative target inspection maneuvers\cite{fourie2014flight}).

\section{GNC strategy}
\begin{figure}
    \centering
    \includegraphics[width=0.8\linewidth]{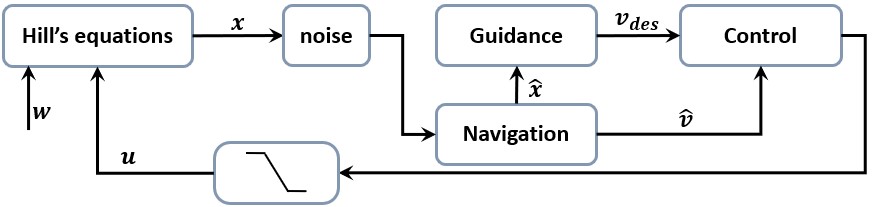}
    \caption{Scheme of the orbital simulator}
    \label{fig:scheme}
\end{figure}

This section presents the proposed GNC strategy to make the spacecraft actually follow the CRO described in the previous section. As briefly described in the Introduction, the guidance function is based on the LGVF method, and takes in input the actual relative position to generate the reference velocity which would drive the spacecraft to the CRO. Then, the velocity feedback control law is presented and, through rigorous Lyapunov analysis, the stability margins are derived based on the bounded actuation power. Additionally, two different state-observers (Luenberger and Levant-based) to perform the navigation function are presented. Finally, the adaptive law which allows for time-varying observer and control gains is discussed. The scheme of the orbital simulator is in Figure \ref{fig:scheme}.

\subsection{Guidance and control strategies}
Autonomous trajectory tracking is based on velocity tracking problem.
The guidance algorithm based on the LGVF approach generates a velocity field based on spacecraft position to track a circular trajectory of radius $R>0$. It is given by
\begin{align}
        u^*&=\delta\left[-x^*(r^2-R^2)+2y^*rR\right],\nonumber\\
        v^*&=\delta\left[-y^*(r^2-R^2)-2x^*rR\right],\label{eq:v_cro}\\
        w^*&=\delta\left[-\lambda z^*r\right]\nonumber,
\end{align}
where $\lambda>0$, and
\begin{align}
        r&=\sqrt{x^{*2}+y^{*2}},\\
        \delta&=\frac{\omega R}{r\sqrt{(r^2+R^2)^2+\lambda z^{*4}}}.
\label{eq:r_delta}
\end{align}
The guidance algorithm is expressed in the trajectory reference frame (denoted with an $^*$) which is rotated of $\pi/3$ aroud the x-axis w.r.t. the LVLH reference frame (in accordance with CRO trajectory definition). For this reason the desired velocity $v_{des}$ is given as
\begin{equation}
     v_{des}=
    Q
    \begin{bmatrix}
        u^*\\
        v^*\\
        w^*
    \end{bmatrix},
\end{equation}
where the matrix $Q$ is given as
\begin{equation}
    Q=
    \begin{bmatrix}
        1 & 0 & 0\\
        0 & \text{cos}(\pi/3) & -\text{sin}(\pi/3)\\
        0 & \text{sin}(\pi/3) & \text{cos}(\pi/3)
    \end{bmatrix}.
\label{eq:Q}\end{equation}
Similarly, transposed matrix of $Q$ is used to determine the spacecraft position in the trajectory reference frame
\begin{equation}
        p^*=Q^Tp
\end{equation}
A feedback controller is designed to track the desired velocity $v_{des}=v_{des}(p)$. In practice, guidance and control input are evaluated using the estimated state of the system (given by the navigation algorithm). Therefore, control input is evaluated as
\begin{equation}
    u=-\text{sat}\left(K\left[\hat{v}-v_{des}(\hat{p})\right]\right),
    \label{eq:control_input}
\end{equation}
where sat$(\cdot)$ is the saturation function, guaranteeing actuators bounded thrust ($|u|\leq U$), and $K>0$ is the constant feedback gain.
Closed-loop system stability is evaluated considering the Lyapunov function
\begin{equation}
    V=\frac{1}{2}\left[v-v_{des}(p)\right]^T\left[v-v_{des}(p)\right].
    \label{eq:lyapunov}
\end{equation}
Assuming the spacecraft to be in proximity of the CRO trajectory ($r\simeq R,\ y\simeq\sqrt{3}z$), the desired velocity is expressed by
\begin{equation}
    \tilde{v}_{des}(p)=
    \begin{bmatrix}
        2\omega z\\
        -\sqrt{3}/2\omega x\\
        -1/2\omega x
    \end{bmatrix}.
    \label{eq:vdes_cro}
\end{equation}
Under this assumption, the time derived of the Lyapunov function expressed in Equation \eqref{eq:lyapunov} is expressed by
\begin{equation}
    \dot{V}=(\dot{x}-2\omega z)(\ddot{x}-2\omega \dot{z})+ (\dot{y}+\sqrt{3}/2\omega x)(\ddot{x}+\sqrt{3}/2\omega \dot{x}) + (\dot{z}+1/2\omega x)(\ddot{z}+1/2\omega \dot{x}).
    \label{eq:V_dot}
\end{equation}
Considering the system dynamics given by Equation \eqref{eq:hill}, and assuming external disturbances and velocity error to be bounded ($||w||_\infty\leq W$ and $||v-v_{des}(p)||_\infty\leq D$), Equation \eqref{eq:V_dot} is maximized by
\begin{equation}
    \dot{V}\leq D(u_x+u_y+u_z)+\left(\frac{\sqrt{3}}{2}+\frac{3}{2}\right)\omega D^2+W D.
\end{equation}
Assuming the estimation error to be small ($\hat{p}\simeq p,\ \hat{v}\simeq v$), the maximum available thrust greater than the maximum disturbance ($U>W$), and $KD\geq U$, then $\dot{V}<0$ if
\begin{equation}
    0\leq D< \frac{6(U-W)}{\omega(3+\sqrt{3})}=\bar{D},
\end{equation}
identifying a bound on the maximum velocity error. This condition is necessary but not sufficient to guarantee asymptotically stability.

Assume the velocity error to be small and comparable with the velocity estimation error $ee_v=\hat{v}-v$ (where $\hat{v}$ is the observer value) and the spacecraft to be in proximity to the CRO, such that Equation \eqref{eq:vdes_cro} is valid. In this case, $v_{des}=\tilde{v}_{des}$ is linear, and Equation \eqref{eq:control_input} is expressed by 
\begin{equation}
    u=-K\left[v-\tilde{v}_{des}(p)\right]-K\left[ee_v-\tilde{v}_{des}(ee_p)\right].
\end{equation}
Under these assumptions, and assuming $||ee_v-v_{des}(ee_p)||_\infty\leq E$, Equation \eqref{eq:V_dot} is maximized by
\begin{equation}
    \dot{V}\leq -KD^2+KDE+\left(\frac{\sqrt{3}}{2}+\frac{3}{2}\right)\omega D^2+W D.
\end{equation}
In this case, $\dot{V}<0$ if
\begin{equation}
    K(D^2-DE)>\left(\frac{\sqrt{3}}{2}+\frac{3}{2}\right)\omega D^2+W D,
\end{equation}
which is valid when
\begin{equation}
    \begin{cases}
        0\leq E <D\\
    K(D^2-DE)>\frac{(\sqrt{3}+3)\omega D^2}{6}+\frac{WD}{2}
    \end{cases}.
    \label{eq:K}
\end{equation}
This relationship identify a direct correlation between the estimation factor $E$ and control gain $K$. Choosing $K$ according to Equation \eqref{eq:K} allows to achieve a trade-off between tracking accuracy and smooth control action.

\subsection{Luenberger Observer}
In this work, a Luenberger Observer is used to estimate the unmeasured states of the spacecraft, i.e. its relative velocity $v$ as in Equation \eqref{eq:hill}. The observer is built as a linear system driven by the available outputs and inputs of the original system as follows:
\begin{equation}\begin{split}
    e_{{x}}&=\Hat{x}-x,\quad e_{{y}}=\Hat{y}-y,\quad e_{{z}}=\Hat{z}-z\\
    \begin{bmatrix}
    \dot{\Hat{x}}\\
    \dot{\Hat{y}}\\
    \dot{\Hat{z}}\\
    \ddot{\Hat{x}}\\
    \ddot{\Hat{y}}\\
    \ddot{\Hat{z}}
\end{bmatrix}&=\underbrace{\begin{bmatrix}
    0 & 0 & 0 & 1 & 0 & 0\\
    0 & 0 & 0 & 0 & 1 & 0\\
    0 & 0 & 0 & 0 & 0 & 1\\
    0 & 0 & 0 & 0 & 0 & 2\omega\\
    0 & -\omega^2 & 0 & 0 & 0 & 0\\
    0 & 0 & 3\omega^2 & -2\omega & 0 & 0
\end{bmatrix}}_{A}\begin{bmatrix}
    {\Hat{x}}\\
    {\Hat{y}}\\
    {\Hat{z}}\\
    \dot{\Hat{x}}\\
    \dot{\Hat{y}}\\
    \dot{\Hat{z}}
\end{bmatrix}+\begin{bmatrix}
    0 & 0 & 0\\
    0 & 0 & 0\\
    0 & 0 & 0\\
    1 & 0 & 0\\
    0 & 1 & 0\\
    0 & 0 & 0\\
\end{bmatrix}\begin{bmatrix}
    u_x\\
    u_y\\
    u_z
\end{bmatrix}-\begin{bmatrix}
    L_{x} & 0 & 0\\
    0 & L_{y} & 0\\
    0 & 0 & L_{z}\\
    L_{\dot{x}} & 0 & 0\\
    0 & L_{\dot{y}} & 0\\
    0 & 0 & L_{\dot{z}}
\end{bmatrix}\begin{bmatrix}
    e_{{x}}\\
    e_{{y}}\\
    e_{{z}}
\end{bmatrix}
\label{eq:Luenb_plant}\end{split}\end{equation}
where $({x},{y},{z})$ are the measured relative positions and $\hat{p}=[\Hat{x},\Hat{y},\Hat{z}]\in\mathbb{R}^3$ are the observer values, so that $ee_p=[e_{{x}},e_{{y}},e_{{z}}]\in\mathbb{R}^3$ denote the observer errors. Also, the time derivatives of the observer values are built in such a way that the observer plant, Equation \eqref{eq:Luenb_plant}, mimics the dynamics system, Equation \eqref{eq:hill}. 
According to Eqs. \eqref{eq:hill} and \eqref{eq:Luenb_plant}, the observer error dynamics is derived as follows:
\begin{equation}
    \begin{bmatrix}
        \dot{e}_x\\
        \dot{e}_y\\
        \dot{e}_z\\
        \ddot{e}_x\\
        \ddot{e}_y\\
        \ddot{e}_z
    \end{bmatrix}=\left(A-\begin{bmatrix}
    L_{x}\\
    L_{y}\\
    L_{z}\\
    L_{\dot{x}}\\
    L_{\dot{y}}\\
    L_{\dot{z}}
\end{bmatrix}\underbrace{\begin{bmatrix}
    1 & 1 & 1 & 0 & 0 & 0 
\end{bmatrix}}_{C}\right)\begin{bmatrix}
        {e}_x\\
        {e}_y\\
        {e}_z\\
        \dot{e}_x\\
        \dot{e}_y\\
        \dot{e}_z
\end{bmatrix}
\label{eq:Luenb_err_dyn}\end{equation}
and the observer gains $L=[L_x,L_y,L_z,L_{\dot{x}},L_{\dot{y}},L_{\dot{z}}]\in\mathbb{R}^6$ are tuned in such a way that the matrix $A_{eq}=A-LC$ is Hurwitz. In particular, the poles of the closed-loop system \eqref{eq:Luenb_err_dyn} are placed considering some desirable properties of the observer, as explained below. First, we neglect both the coupling terms in the matrix $A$ and the control input $u$, and we consider null observer estimation at $t=0$. Consequently, for each axis the observer dynamics \eqref{eq:Luenb_plant} is described by a second order single input single output (SISO) system. In the following, the derivation of the observer gains is specified for the $x-$axis, and it is equivalent for both $y$ and $z$ axis. 
The resulting SISO system and relevant transfer function are as follows:
\begin{equation}\begin{split}
    \Ddot{\Hat{x}}=L_x\left(x-\Hat{x}\right)-L_{\dot{x}}\dot{\Hat{x}},\quad    {\Hat{x}}=\frac{1}{1+2\frac{\zeta}{\omega_n}s+\frac{1}{\omega_n^2}}x\quad\Rightarrow\quad L_x=\omega_n^2,~L_{\dot{x}}=2\zeta\omega_n,
\end{split}\end{equation}
where $\zeta$ and $\omega_n$ are derived from the specifications on the maximum overshoot $\overline{\hat{x}}$ and the rise time $\overline{t}_s$ of the desired response:
$$
\zeta=-\frac{\text{ln } \overline{\hat{x}}}{\sqrt{\pi+\text{ln } \overline{\hat{x}}^2}},\quad \omega_n=\frac{\pi}{\overline{t}_s\sqrt{1-\zeta^2}}.
$$

\subsection{Levant Observer}
This section introduces the Levant extended state observer (LeO) to estimate both the unmeasured relative velocity of the spacecraft and the 
total disturbances affecting the position dynamics \eqref{eq:hill} basing on the Levant's robust differentiator.  
\cite{doi:10.1080/0020717031000099029} At this aim, the disturbances ${w}_d=[w_{d,x}, w_{d,y}, w_{d,z}]^T$ are treated as extended states and an extended system is obtained as follows from Equation \eqref{eq:hill}:
\begin{equation}\begin{split}
    \dot{{\xi}}_1&={{\xi}}_2\\
    \dot{{\xi}}_2&={{\xi}}_3+f\left({{\xi}}_1,{{\xi}}_2\right)+{u}\\
    \dot{{\xi}}_3&={{\xi}}_4\\
    \dot{{\xi}}_4&=\ddot{{w}}_d
\end{split}\label{eq:ExtSys},\end{equation}
where $\xi_1\coloneqq p$, $\xi_2\coloneqq v$, $\xi_3=w_d$, $f\left({\xi}_1,{{\xi}}_2\right)=[2\omega\Dot{z}, -\omega^2y, 3\omega^2z-2\omega\dot{x}]^T$, and ${u}=[u_x,u_y,u_z]^T$. Then, the extended state observer is designed as follows:
\begin{equation}\begin{split}
    \dot{\hat{{\xi}}}_1&=\hat{{\xi}}_2-8.6\left(k_4L\right)^\frac{1}{4}\left|{e_\xi}_1\right|^\frac{3}{4}\text{sign}({e_\xi}_1)\\
    \dot{\hat{{\xi}}}_2&=\hat{{\xi}}_3-21\left(k_4L\right)^\frac{2}{4}\left|{e_\xi}_1\right|^\frac{2}{4}\text{sign}({e_\xi}_1)+f\left({\hat{{\xi}}}_1,{\hat{{\xi}}}_2\right)+{u}\\
    \dot{\hat{{\xi}}}_3&=\hat{{\xi}}_4-16.25\left(k_4L\right)^\frac{3}{4}\left|{e_\xi}_1\right|^\frac{1}{4}\text{sign}({e_\xi}_1)\\
    \dot{\hat{{\xi}}}_4&=-k_4L\text{sign}({e_\xi}_1)
\end{split}\label{eq:Levant_plant}\end{equation}
where the observer errors $e_{\xi_1}=\hat{{\xi}}_1-{{\xi}}_1$ is obtained as the difference between the observer values $\hat{{\xi}}_1$ of the measured position ${\xi}_1$. Since Equation \eqref{eq:Levant_plant} is discontinuous, its solutions are understood in the sense of Filippov \cite{filippov2013differential}. Then, the observer errors dynamics is obtained from \eqref{eq:ExtSys} and \eqref{eq:Levant_plant} as follows:
\begin{equation}\begin{split}
    {{e}_\xi}_2&=\dot{\hat{{\xi}}}_1-\dot{{{\xi}}}_1=\hat{{\xi}}_2-{{\xi}}_2-8.6\left(k_4L\right)^\frac{1}{4}\left|{e_\xi}_1\right|^\frac{3}{4}\text{sign}({e_\xi}_1)\\
    {{e}_\xi}_3&=\dot{\hat{{\xi}}}_2-\dot{{{\xi}}}_2=\hat{{\xi}}_3-{{\xi}}_3+f\left({\hat{{\xi}}}_1-{{{\xi}}}_1,{\hat{{\xi}}}_2-{{{\xi}}}_2\right)-21\left(k_4L\right)^\frac{2}{4}\left|{e_\xi}_1\right|^\frac{2}{4}\text{sign}({e_\xi}_1)\\
    {{e}_\xi}_4&=\dot{\hat{{\xi}}}_3-\dot{{{\xi}}}_3=\hat{{\xi}}_4-{{\xi}}_4-16.25\left(k_4L\right)^\frac{3}{4}\left|{e_\xi}_1\right|^\frac{1}{4}\text{sign}({e_\xi}_1)\\
    {{e}_\xi}_5&=\dot{\hat{{\xi}}}_4-\dot{{{\xi}}}_4=-k_4L\text{sign}({e_\xi}_1)-\ddot{{w}}_d
\end{split}\end{equation}
where $f({\hat{{\xi}}}_1,\hat{{{\xi}}}_2)-f({{{\xi}}}_1,{{{\xi}}}_2)=f({\hat{{\xi}}}_1-{{{\xi}}}_1,{\hat{{\xi}}}_2-{{{\xi}}}_2)$ due to the linearity of $f$. Furthermore, $L>\|\ddot{{w}}_d\|_\infty$ and $k_4>L$ are selected in order to guarantee that the errors ${e_\xi}_i$, $i=1,\dots,5$ converge to zero in finite-time with no noise in the loop. \cite{8485770} Consequently, $\hat{{\xi}}_3$ gives the estimation of the disturbance ${w}_d$ and $\hat{{\xi}}_2$ provides the estimation of the spacecraft velocities $v=[\dot{x},\dot{y},\dot{z}]^T$.

\subsection{Adaptive Law}
Observer gains design struggle to achieve fast convergence and low noise sensitivity. High observers gains ensure quick convergence, but also cause low estimation accuracy and more sensitivity to sensor noise. Reducing the observer gains (within convergence bounds) improves estimation accuracy, but compromise convergence speed. 
For this reason, an adaptive law is proposed to achieve both fast convergence and low noise sensitivity. The adaptive parameter time derivative changes according to a linear function\cite{lahana2023comparison}, based on the current estimated state. It is given by
\begin{equation}
\dot{\delta}=\text{Proj}_{[\underline{\delta},\bar{\delta}]}\left(G||\hat{v}-\tilde{v}_{des}(\hat{p})||_\infty-\gamma\delta\right),
\label{eq:adaptive}
\end{equation}
where $G,\gamma>0$ are constant parameters designed in order to guarantee closed-loop system stability, $\underline{\delta}$ guarantees observer convergence and stability, and $\bar{\delta}=1$. 
The design of $G$ and $\gamma$, is based on definition of the CRO tracking bound ($||\hat{v}-\tilde{v}_{des}(\hat{p})||_\infty\leq\Gamma<\bar{D}$), and considering $\gamma=G\Gamma$, imposing
\begin{equation}
    \dot{\delta}\leq0 \iff ||\hat{v}-\tilde{v}_{des}(\hat{p})||_\infty\leq\Gamma.
\end{equation}
Then, the decreasing of $\delta$ is bounded considering 
\begin{equation}
    \dot{\delta}\geq-\Delta \iff \gamma\leq\Delta,
\end{equation}
with $\Delta>0$. The parameter $\underline{\delta}\leq\delta\leq\bar{\delta}$ is used to regulate observer gains. Assuming the reactive gain design to guarantee robust observer design, it is reasonable to consider small estimation errors when the spacecraft is in proximity of the CRO trajectory. Under this assumption, if $\Gamma$ is designed accordingly, the observer gains can be reduced to improve estimation accuracy. While LeO convergence can be guaranteed by $\underline{\delta}L>\|\ddot{w}_d\|_\infty$, similar consideration cannot be derived for LO, but the equivalent matrix negative eigenvalues needs to be verified for $A_{eq}=(A-\underline{\delta}LC)$.
Reducing the estimation error allows to reduce also the feedback gain $K$ to avoid high frequency control input, considering the feedback controller
\begin{equation}
    u=-\delta K\left[\hat{v}-v_{des}(\hat{p})\right],
    \label{eq:adaptive_u}
\end{equation}
where $\underline{\delta}K$ needs to satisfy closed-loop system stability condition of Equation \eqref{eq:K}.

\section{Simulations}
This section discussed the results of the numerical simulations. As discussed above, the simulation scenario involves a spacecraft in relative motion around a target along CRO trajectory, as described by Equation \eqref{eq:CRO}. 
The LGVF-based guidance algorithm is implemented according to Eqs. \eqref{eq:v_cro}-\eqref{eq:Q}, and gives the desired velocity to track the objective CRO. The guidance is combined with the velocity feedback controller described by Equation \eqref{eq:control_input}, while sensors are modeled considering realistic vision-based measurements of relative position (cm accuracy). Then, LO and LeO are implemented to estimate spacecraft velocity according to Eqs. \eqref{eq:Luenb_plant} and \eqref{eq:Levant_plant}, respectively.

\begin{figure}[]
	\centering\includegraphics[width=0.49\textwidth]{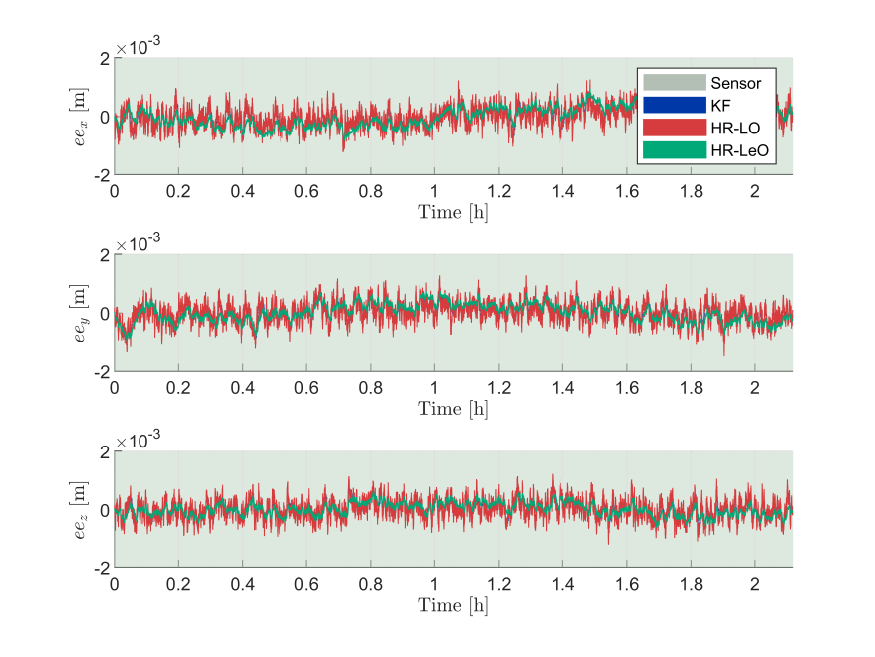}
 \includegraphics[width=0.49\textwidth]{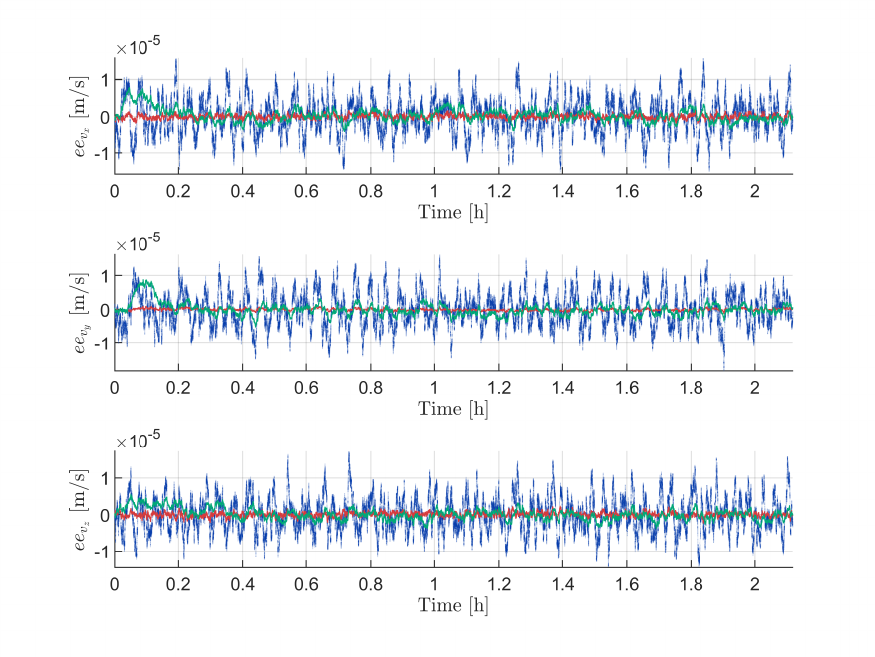}
	\caption{Comparison of Kalman Filter, Luenberger and Levant observers techniques. Position (left) and velocity (right) estimation errors with nominal initial conditions.}
	\label{fig:nominal}
\end{figure}

First, observers performance are evaluated in open-loop and uncontrolled motion ($u=0$) with the spacecraft moving on the CRO. In Figure \ref{fig:nominal}, the performance of High Resolution (HR) observers gain design is compared with Kalman Filter\cite{chen2011kalman} performance. While position estimation accuracy is comparable for the three methods, HR gains allow the observers to outclass Kalman Filter in velocity estimation. 

\begin{figure}[htb]
	\centering
    \includegraphics[width=0.6\textwidth]{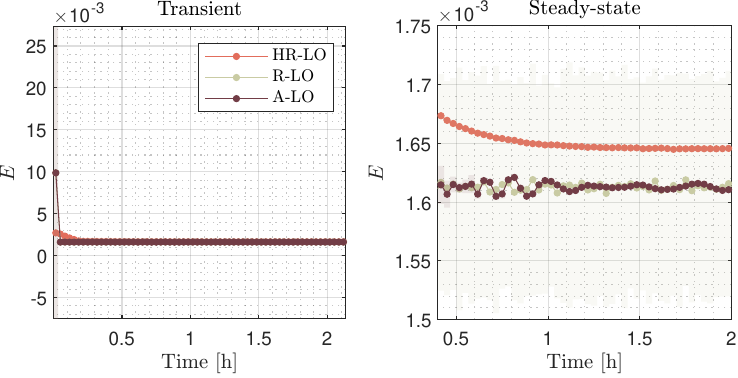}
	\caption{Estimation factor variation with non-nominal initial conditions: Luenberger observer.}
	\label{fig:comparison1}
\end{figure}

\begin{figure}[htb]
	\centering
    \includegraphics[width=0.6\textwidth]{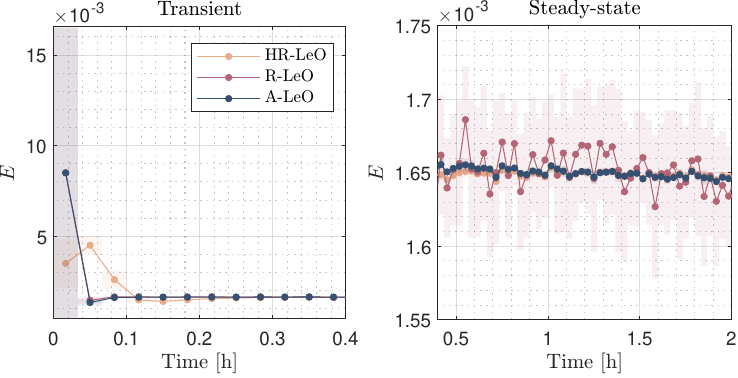}
	\caption{Estimation factor variation with non-nominal initial conditions: Levant observer.}
	\label{fig:comparison2}
\end{figure}

\begin{figure}[htb]
	\centering
    \includegraphics[width=0.6\textwidth]{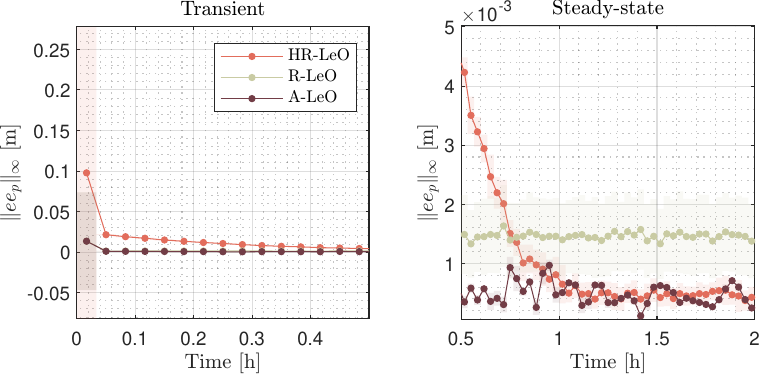}\\
    \includegraphics[width=0.6\textwidth]{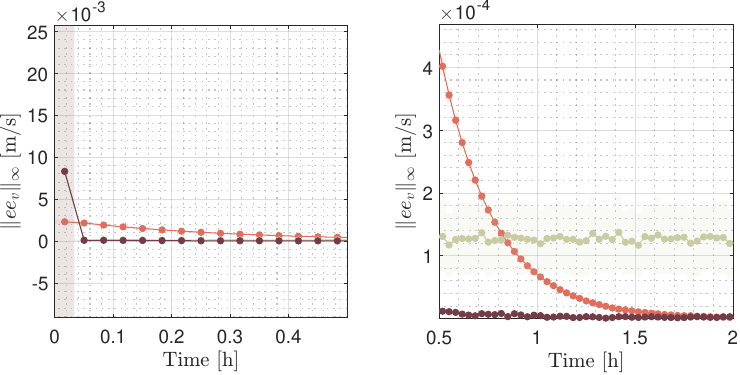}
	\caption{LO estimation performance with non-nominal initial conditions: position and velocity estimation errors.}
	\label{fig:comparison3}
\end{figure}

Then, Figures \ref{fig:comparison1}-\ref{fig:comparison3} compares the results obtained with: (i) Adaptive (A) gains, constant observer gains with HR tuning, and (iii) constant observer gains with Reactive (R) tuning. The comparison is performed considering the estimation factor $E$ with both LO (Fig.\ref{fig:comparison1}) and LeO (Fig.\ref{fig:comparison2}) under non-nominal initial conditions (errors of m for position and mm/s for velocity).
Results show that the HR design lacks of fast convergence properties in case of big initial estimation errors. 
Instead, R gains design achieves fast convergence but compromises estimation accuracy. The A strategy (Equation \eqref{eq:adaptive}) exploits the best characteristics of R and HR designs, providing both fast convergence and low noise sensitivity. 
For completeness, in Figure \ref{fig:comparison3} are also shown position and velocity estimation errors.
The stability properties of the A-LO are affected by the time-varying behavior of $\delta$. Indeed, the design of $L$ is obtained to guarantee fast convergence (R-LO design).
However, for $\delta<1$, observer gains are reduced according to $\delta L$, so that the real part of the eigenvalues of $A-LC$ shift towards the zero value as $\delta$ decreases. 
In particular, for $\delta=\underline{\delta}$ some eigenvalues have real part close to zero, as shown below, and shift to imaginary values which degrades the convergence properties of the observer.
\begin{equation}
    A_{eq}=A-LC\to\lambda=[-1,\ -1,\ -1,\ -0.1,\ -0.1,\ -0.1]
\end{equation}
\begin{equation}
    A_{eq}=A-\underline{\delta}LC\to\lambda=10^{-2}[-0.01 \pm 0.39i,\ -0.001 \pm 0.23i,\ -0.001 \pm 0.33i]
\end{equation}
However, A-LeO does not suffer of the same problem, since asymptotically convergence is guaranteed by $\underline{\delta}L>\|\ddot{w}_d\|_\infty$.

\begin{figure}[htb]
	\centering
    \includegraphics[width=0.6\textwidth]{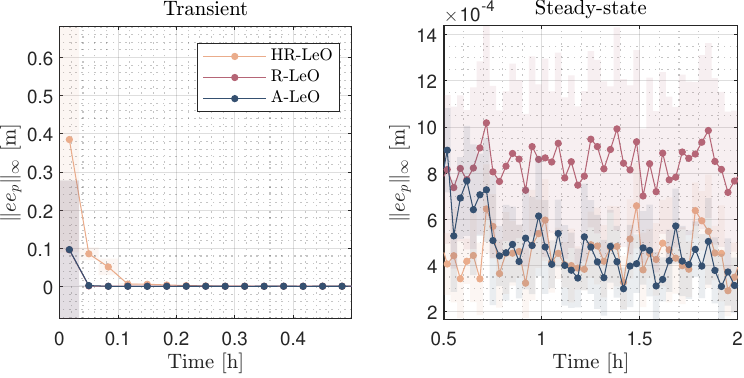}\\
    \includegraphics[width=0.6\textwidth]{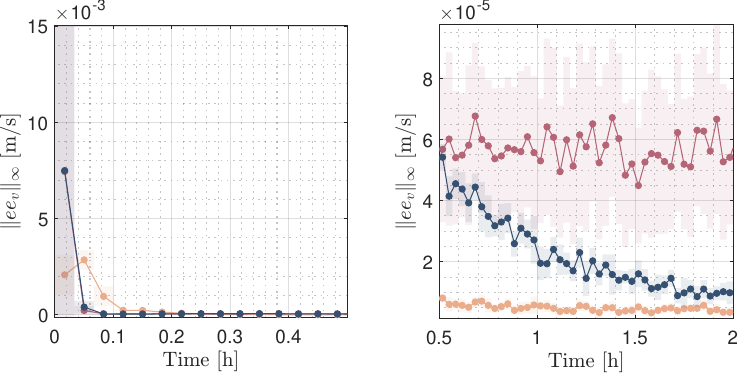}
	\caption{LeO estimation performance with non-nominal initial conditions: position and velocity estimation errors.}
	\label{fig:comparison4}
\end{figure}

\begin{figure}[]
	\centering
 \includegraphics[width=0.4\textwidth]{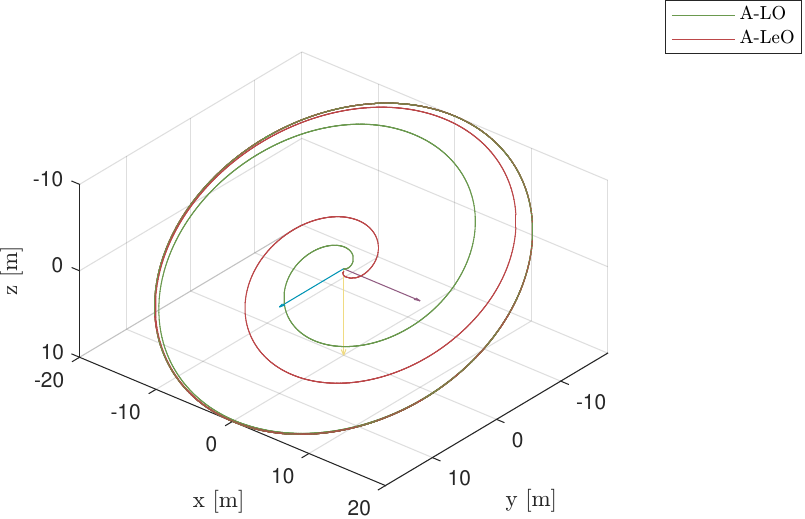}
 \includegraphics[width=0.4\textwidth]{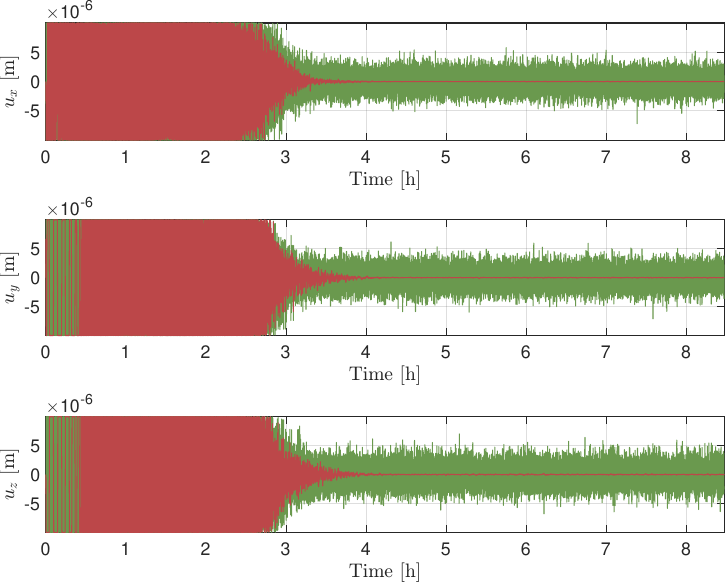}
 \caption{Spacecraft deploying in CRO: trajectory and control input}
	\label{fig:trajectory}
\end{figure}
\begin{figure}[]
	\centering
 \includegraphics[width=0.7\textwidth]{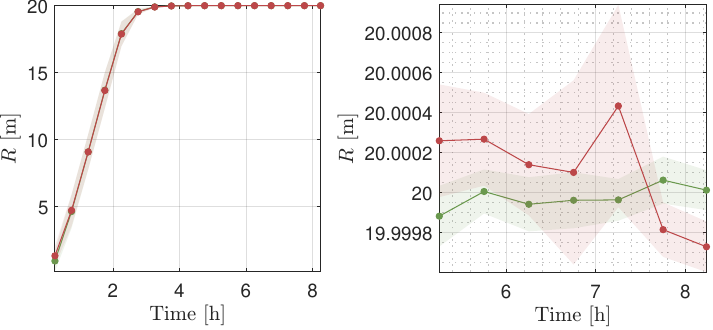}
 \caption{Spacecraft deploying in CRO: trajectory tracking accuracy}
	\label{fig:radius}
\end{figure}

Closed-loop system performance are compared considering the observers and controller (Equation \eqref{eq:adaptive_u}) adaptive design in spacecraft deploying mission in CRO. Closed-loop system tracking performance are compared in Figures \ref{fig:trajectory} and \ref{fig:radius}. A-LO is slightly more accurate for trajectory tracking than A-LeO, but this is paid with high switching frequency of the control input (right plot of Figure \ref{fig:trajectory}). A-LeO allows to achieve accurate state estimation, minimizing control input switching, and reaching mm-accuracy trajectory tracking. This trade-off underscores the importance of selecting the appropriate adaptive design based on mission requirements, where A-LO may be preferred for scenarios prioritizing tracking accuracy, and A-LeO for those emphasizing long-term system efficiency and stability.



\section{Conclusions}
Autonomous space operations are critical for the success of deep space exploration and satellite maintenance, where communication delays and the absence of Earth-based signals necessitate self-reliant GNC systems. This study presented an autonomous GNC strategy for spacecraft performing proximity operations, specifically focusing on the task of maintaining a CRO trajectory around a target. The proposed solution employed the LGVF method to generate the desired trajectory and a velocity feedback controller to track the CRO, but the performance of this system is significantly influenced by the accuracy of onboard sensor data.
To address the challenges of noise and estimation errors in real-time spacecraft localization, LO and LeO are compared wiht Kalman Filter. While LO and LeO methods offer strong estimation capabilities, they struggle to balance fast convergence with sensor noise resilience. To overcome these limitations, this work introduced an adaptive observer strategy with time-varying gains. By adjusting the observer gains based on trajectory tracking error, the proposed method achieved fast convergence during large deviations and reduced noise sensitivity as the system approached the desired trajectory.
Closed-loop system performance are evaluated by means of stability analysis and numerical simulations. Results validate the effectiveness of the adaptive observer in mitigating the impact of sensor noise and enhancing trajectory control. This advancement is particularly relevant for critical operations such as proximity formation flight and uncooperative target inspection, where precise navigation and control are essential for avoiding collisions and ensuring mission success.  
In conclusion, the comparative analysis of the adaptive observer designs highlights the inherent trade-offs in spacecraft deploying missions in CRO. The A-LO design offers slightly higher precision in trajectory tracking but requires frequent control input switching, which may impact system efficiency and hardware longevity. Conversely, the A-LeO design achieves a balanced performance, delivering accurate state estimation and trajectory tracking while significantly reducing control input switching, thereby enhancing overall system robustness and sustainability. These findings underscore the importance of tailoring the adaptive design approach to specific mission objectives, with A-LO being suitable for missions demanding maximum tracking accuracy, and A-LeO better suited for scenarios prioritizing efficiency and stability in the long term.

\bibliographystyle{AAS_publication}   
\bibliography{references}   

\end{document}